\pdfoutput=1

\documentclass[11pt]{article}

\usepackage{acl}

\usepackage{times}
\usepackage{latexsym}
\usepackage{graphicx}
\usepackage{amsmath}
\usepackage{amssymb}
\usepackage{xspace}
\newcommand{\mName}{DualDe\xspace}

\usepackage[T1]{fontenc}

\usepackage[utf8]{inputenc}

\usepackage{microtype}

\usepackage{inconsolata}

\usepackage{graphicx}
\usepackage{amsmath}
\title{A Dual-Module Denoising Approach with Curriculum Learning for Enhancing Multimodal Aspect-Based Sentiment Analysis}

\author{Nguyen Van Doan, Dat Tran Nguyen, Cam-Van Thi Nguyen\thanks{Corresponding author.
Cam-Van Thi Nguyen was funded by the Master, PhD Scholarship Programme of Vingroup Innovation Foundation (VINIF), code VINIF.2023.TS147.} \\
        Faculty of Information Technology \\
        VNU University of Engineering and Technology \\
        \texttt{\{21020111, 21020011, vanntc\}@vnu.edu.vn}}

\begin{document}
\maketitle
\begin{abstract}
Multimodal Aspect-Based Sentiment Analysis (MABSA) combines text and images to perform sentiment analysis but often struggles with irrelevant or misleading visual information. Existing methodologies typically address either sentence-image denoising or aspect-image denoising but fail to comprehensively tackle both types of noise. To address these limitations, we propose \textbf{\mName{}}, a novel approach comprising two distinct components: the \textit{Hybrid Curriculum Denoising Module} (HCD) and the \textit{Aspect-Enhance Denoising Module} (AED). The HCD module enhances sentence-image denoising by incorporating a flexible curriculum learning strategy that prioritizes training on clean data. Concurrently, the AED module mitigates aspect-image noise through an aspect-guided attention mechanism that filters out noisy visual regions which unrelated to the specific aspects of interest. Our approach demonstrates effectiveness in addressing both sentence-image and aspect-image noise, as evidenced by experimental evaluations on benchmark datasets.  
\end{abstract}

\section{Introduction}

Sentiment analysis is a fundamental task in natural language processing (NLP) \cite{liu2012sentiment}, which seeks to uncover and interpret the opinions, attitudes, and emotions embedded in user-generated content. Multimodal Aspect-Based Sentiment Analysis (MABSA) extends this analysis by combining textual and visual modalities to achieve a deeper and more comprehensive understanding of sentiment. MABSA is typically organized into three principal subtasks: Multimodal Aspect Term Extraction (MATE), which focuses on the identification and extraction of aspect-specific terms from text \cite{wu2020multimodal}; Multimodal Aspect-Oriented Sentiment Classification (MASC), which involves classifying the sentiment associated with each aspect term into categories such as positive, neutral, or negative \cite{yu2019adapting}; and Joint Multimodal Aspect-Sentiment Analysis (JMASA), which concurrently addresses aspect extraction and sentiment classification to provide a unified analysis of both aspects and sentiments \cite{ju2021joint}.

In real-world scenarios, not all images are relevant to the accompanying text; some even mislead the contextual and emotional understanding of the sentence. For images that are related to the text, not all visual blocks in the image are closely tied to the aspect; in fact, there are often blocks that introduce noise.
\begin{figure}[!t]
    \centering
    \includegraphics[width=1\linewidth]{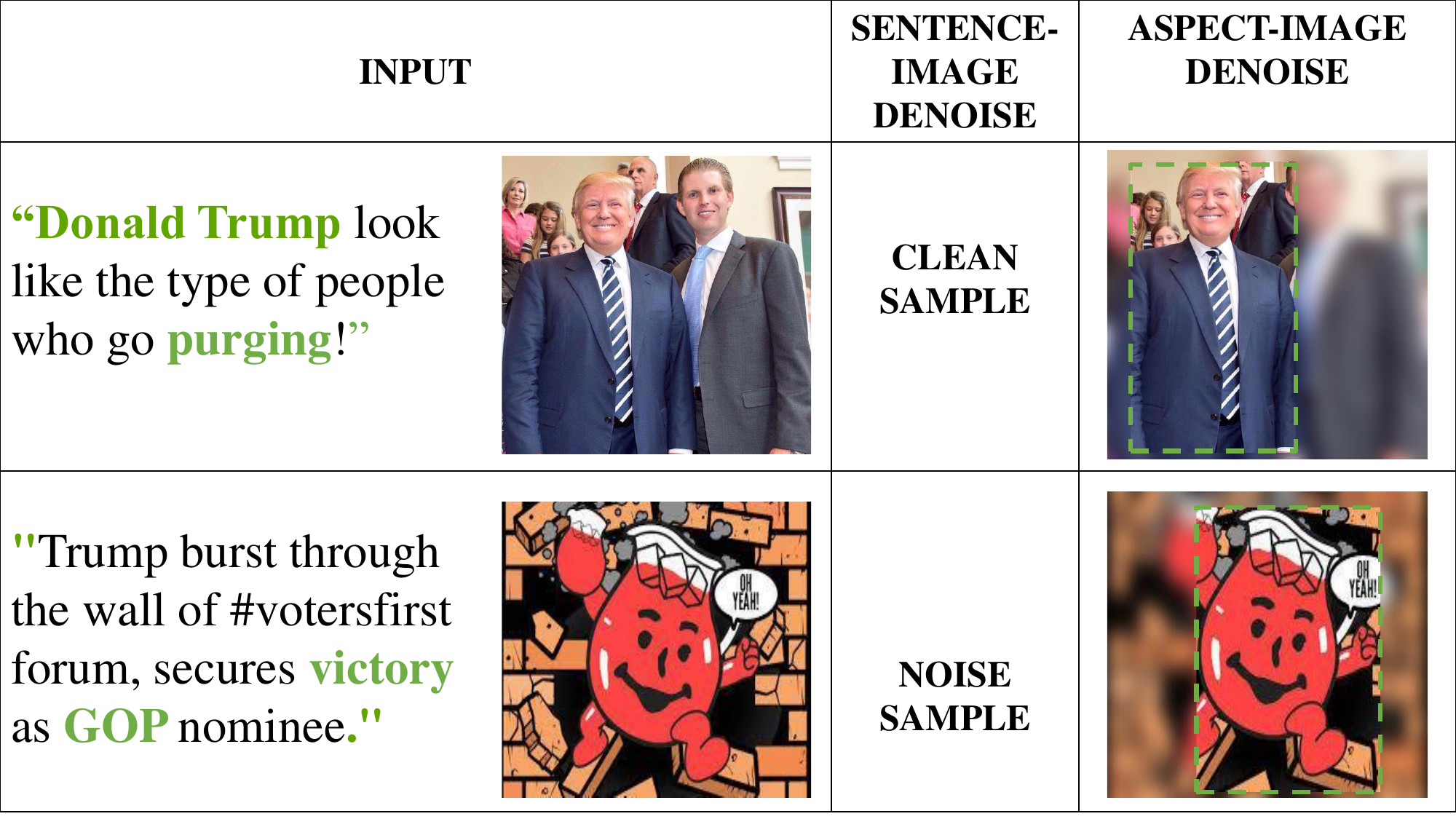}
    \caption{Illustration of Sentence-Image Denoising and Aspect-Image Denoising. Sentence-Image Denoising classifies an image as clean if it is relevant to the overall sentence meaning. Aspect-Image Denoising identifies regions as noise (e.g., blurred areas) when they lack strong relevance to any specific aspect.}
    \label{fig:noise_example}
\end{figure}
In real-world scenarios, images accompanying text may not always be relevant and can sometimes mislead the interpretation of the sentence's context and emotion. Even when images are relevant, not all visual elements are tied to the aspect of interest, often introducing noise. To address these challenges, existing methods focus on either sentence-image or aspect-image denoising. Approaches such as those by \cite{ju2021joint} and \cite{sun2021rpbert} utilize text-image relation detection to filter out non-contributory visual information but may miss significant details in images deemed irrelevant. \cite{zhao2023m2df} address this with Curriculum Learning, progressively exposing the model to noisy images; however, their fixed noise metric limits flexibility. On the other hand, methods like those by \cite{zhang2021multimodal} and \cite{yu2022targeted} concentrate on the interaction between visual objects and specific words, while \cite{zhou2023aom} use an aspect-aware attention module for fine-grained alignment. Despite their advantages, these methods often neglect the importance of sentence-image denoising, as illustrated in Figure \ref{fig:noise_example}.

In this paper, we propose \textbf{\mName{}}, an advanced approach designed to comprehensively address both sentence-image and aspect-image noise. \mName{} integrates two principal components: the Hybrid Curriculum Denoising Module (HCD) and the Aspect-Enhance Denoising Module (AED). The Hybrid Curriculum Denoising Module advances sentence-image denoising by implementing a flexible curriculum learning approach that dynamically adjusts noise metrics based on both model performance and pre-defined standards, thereby enhancing adaptability. The Aspect-Enhance Denoising Module (AED) utilizes an aspect-guided attention mechanism to selectively filter out irrelevant visual regions and textual tokens related to each specific aspect, thereby improving image-text alignment. Our contributions are summarized as follows:
\begin{itemize}
    \item To the best of our knowledge, we are the first to present a model, \mName{}, that concurrently addresses both sentence-image and aspect-image noise.
    \item We introduce the Hybrid Curriculum Denoising Module(HCD), which effectively balances generalization and adaptability within the training framework.
    \item We demonstrate the effectiveness of our approach through extensive experiments on the Twitter-15 and Twitter-17 datasets. 
\end{itemize}

\section{Related Work}
\subsection{Multimodal Aspect-based Sentiment Analysis}

With the proliferation of social media, where posts frequently encompass multiple modalities such as text and images, there has been considerable interest in utilizing multimodal approaches to analyze aspects and sentiments in user-generated content \cite{cai2019multi}. The Multimodal Aspect-Based Sentiment Analysis (MABSA) task is typically segmented into three core subtasks: Multimodal Aspect Term Extraction (MATE) \cite{wu2020multimodal}, which focuses on identifying aspect terms within text; Multimodal Aspect-Oriented Sentiment Classification (MASC) \cite{yu2019adapting}, which classifies the sentiment associated with each aspect term; and Joint Multimodal Aspect-Sentiment Analysis (JMASA) \cite{ju2021joint}, which integrates MATE and MASC by concurrently extracting aspect terms and predicting their associated sentiments.

With the prevalence of noisy images in multimodal data, several methods have been proposed to address this issue. \citet{ju2021joint} and \citet{sun2021rpbert} address the issue of noisy images by incorporating an auxiliary cross-modal relation detection module that filters and retains only those images that genuinely contribute to the text's meaning. \citet{ling2022vision} propose a Vision-Language Pre-training architecture specifically for MABSA, which enhances cross-modal alignment between text and visual elements, thereby mitigating the impact of noisy visual blocks. Meanwhile, \citet{zhang2021multimodal} and \citet{yu2022targeted} focus on eliminating noise by disregarding image regions without visual objects and concentrating solely on regions containing relevant visual elements and their interaction with text. \citet{zhou2023aom} propose an aspect-aware attention module that enhances image-text alignment by weighting tokens according to their relevance to the aspect, thereby effectively reducing aspect-image noise.

\subsection{Curriculum Learning}
Curriculum Learning (CL), introduced by \citet{bengio2009curriculum}, is a machine learning strategy that mimics human learning by starting with simpler concepts and progressively tackling more complex ones. CL has shown benefits across various tasks \citep{wang2019dynamically, lu2021exploiting, platanios2019competence, nguyen-etal-2024-curriculum} and has been effective in mitigating noisy images in the Multimodal Aspect-Based Sentiment Analysis (MABSA) task \citep{zhao2023m2df}. While \citet{zhao2023m2df} utilize CL to progressively expose the model to noisy images, starting from cleaner data to address sentence-image noise, they do not account for aspect-image noise. In this paper, we extend this concept by proposing the Hybrid Curriculum Denoising Module (HCD), specifically designed to reduce sentence-image noise and enhance overall performance.
\begin{figure*}[t]
    \centering
    \includegraphics[width=0.9\textwidth]{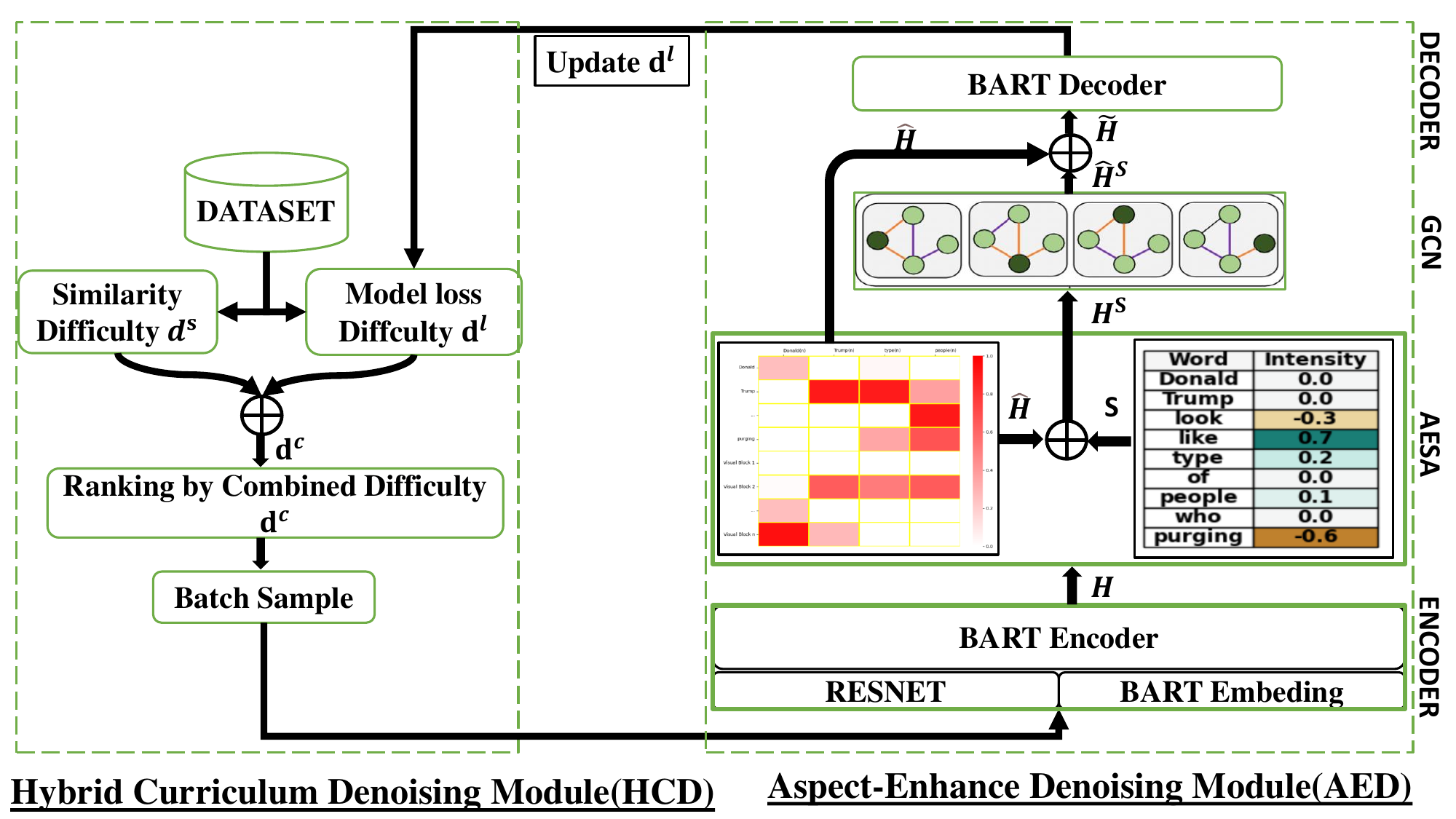}
    \caption{Model Overview}
    \label{fig:overview}
\end{figure*}
\section{Methodology}
Our model comprises two main modules: (1) the Hybrid Curriculum Denoising Module (HCD) and (2) the Aspect-Enhanced Denoising Module (AED). The Aspect-Enhanced Denoising Module (AED) is constructed on a BART-based architecture and incorporates two sub-components situated between the encoder and decoder: Aspect-Based Enhanced Sentic Attention (AESA) and Graph Convolutional Network (GCN). An overview of the architecture is illustrated in Figure \ref{fig:overview}.

\textbf{Task Definition.} In this task, given a tweet with an image \( I \) and a sentence \( T \) consisting of \( m \) words \( T = \{t_1, t_2, \dots, t_m\} \), the objective is to generate an output sequence \( Z = [b_{begin}^1, b_{end}^1, p_1, \dots, b_{begin}^m, b_{end}^m, p_m] \). Each tuple \( [b_{begin}^i, b_{end}^i, p_i] \) represents the \( i \)-th aspect, where \( b_{begin}^i \) and \( b_{end}^i \) denote the starting and ending positions of the aspect, and \( p_i \) indicates its sentiment polarity (Positive, Negative, or Neutral). Aspects can span multiple words, and a single sentence may include multiple aspects, each with different sentiment polarities.

\textbf{Feature Extractor.} We pre-trained BART \cite{lewis2019bart} model for embeddings word and ResNet \cite{chen2014deepsentibank} for embeddings image. The formatted output is \( I = \{\texttt{<img>} i_1 \texttt{</img>}, \dots, \texttt{<img>} i_m \texttt{</img>} \} \) and \( T = \{\texttt{<bos>} t_1 \texttt{<eos>}, \dots, \texttt{<bos>} t_n \texttt{<eos>} \} \) where $m$ is the number of image features extracted by Resnet (surround by \texttt{<img>}...\texttt{</img>}), $n$ is the number of text features(surround by \texttt{<bos>}...\texttt{<eos>}). These features are combined into a sequence \( X \), which is then used as the input for the BART encoder.

The encoder generates multimodal hidden states \( H = \{h_0^I, h_1^I, \dots, h_m^I, h_0^T, h_1^T, \dots, h_n^T \} \), where \( h_i^I \) represents the feature of the \( i \)-th visual block from the image \( I \), and \( h_j^T \) represents the feature of the \( j \)-th word from the sentence \( T \), with \( m \) visual blocks and \( n \) words in total.

\subsection{Hybrid Curriculum Denoising Module (HCD)}
This HCD module employs a flexible training strategy that adapts to varying levels of image noise, starting with cleaner data and progressively incorporating noisier examples. By integrating dynamic noise metrics from both model predictions and predefined standards, this module enhances the model's ability to mitigate sentence-image noise effectively.
\subsubsection{Similarity Difficulty Metric}
As depicted in Figure \ref{fig:noise_example}, when a sentence is paired with images that closely align with its content, it enhances the comprehension of the sentence's meaning and sentiment. Consequently, the degree of similarity between the text and accompanying images can be considered an indicator of learning difficulty: greater similarity suggests an easier learning process, whereas lower similarity indicates increased difficulty. The similarity score is computed as follows:
\begin{equation}
S_{(X_i^T, Y_i^I)} = \cos(X_i^T, Y_i^I)
\end{equation}
where \( S \) is the similarity score calculated by the cosine function \(\cos(\cdot)\), \(X_i^T\) and \(Y_i^I\) represent the textual and visual features, respectively, obtained through the text and image encoders of the pre-trained CLIP model \cite{radford2021learning}.

Subsequently, we define and normalize the difficulty at the sentence level of i-th sample as follows:

\begin{equation}
d^s_i = 1.0 - \frac{\text{S}_{(X^T_i, Y^I_i)}}{\max\limits_{1 \leq k \leq N} \text{S}_{(X^T_k, Y^I_k)}},
\end{equation}

where $N$ is length of train dataset, \(d^s_i\) is normalized within the range \([0.0, 1.0]\). A lower value of \(d^s_i\) indicates that the data is likely to be easier to learn or predict accurately and will therefore be prioritized in the learning process.

\subsubsection{Model loss Diffculty Metric}
The individual loss function for each data sample in a sequential model can be expressed as:
\begin{equation}
L_i = - \sum_{t=1}^{O} \log P(y_t \mid Y_{<t}, X_i)
\end{equation}
where \( L_i \) represents the loss for the \( i \)-th data sample, \( X_i \) is the input for that sample, and \( O \) is the sequence length. \( y_t \) denotes the word or character at time step \( t \), and \( Y_{<t} \) represents all preceding words or characters. \( P(y_t \mid Y_{<t}, X_i) \) is the probability predicted by the model for the word \( y_t \) given the context \( Y_{<t} \) and input \( X_i \). 

After that, we normalized this difficulty score of i-th sample to $[0.0,1.0]$ by following formula:
\begin{equation}
d^l_i = \frac{L_i}{\max\limits_{1 \leq j \leq N} L_j}   
\end{equation}
where $N$ is length of train dataset.

Since the difficulty a batch sample (based on the loss metric) is entirely dependent on the model's state, we update the loss metric at each epoch to ensure accurate evaluation.

\subsubsection{Comprehensive Difficulty Metric}
The difficulty metric \(d^s_i\) is a predefined metric that remains constant throughout the training process. Conversely, the difficulty metric \(d^l_i\) is based on the model's current learning state and changes at each epoch. To balance the generalization of \(d^s_i\) and the adaptability of \(d^l_i\) in training schedules, we propose a new composite difficulty metric  \(d^c_i\) for i-th sample, defined as:
\begin{equation}
d^c_i = \alpha \cdot d^l_i + (1 - \alpha) \cdot d^s_i
\end{equation}
where \(\alpha\) is a weighting factor that balances the contribution of \(d^l_i\) and \(d^s_i\). Empirical results indicate that setting \(\alpha = 0.8\) yields optimal performance.

\subsubsection{Curriculum Training}
\begin{figure}
    \centering
    \includegraphics[width=1\linewidth]{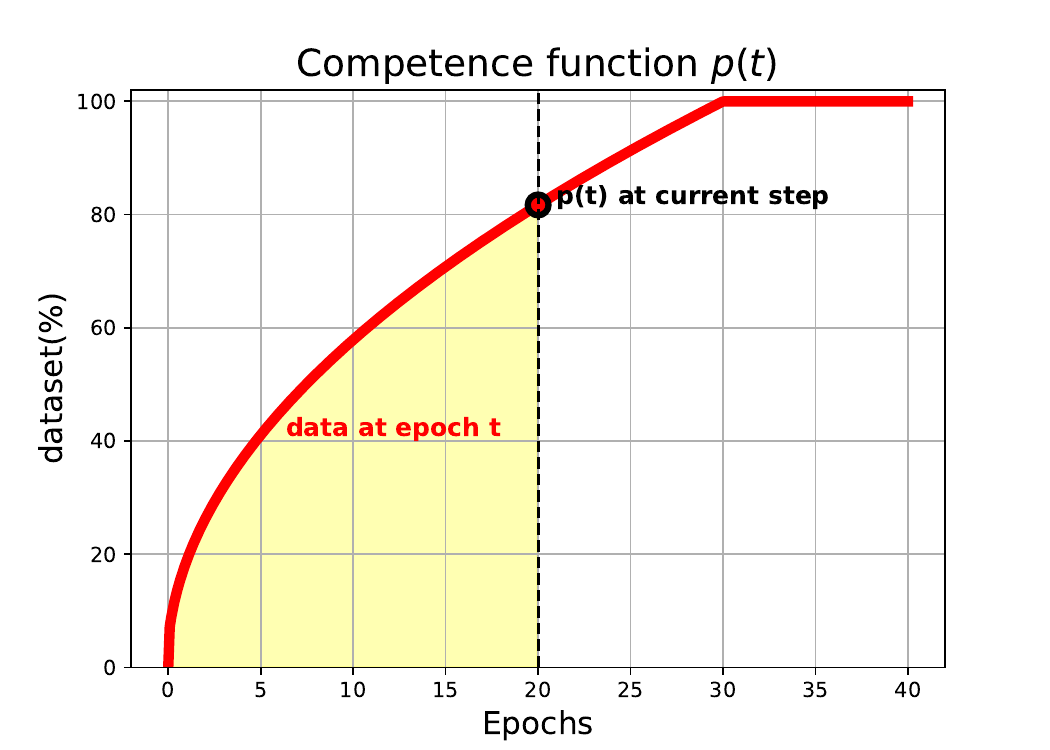}
    \caption{Illustrate the curve of the competence function \(p(t)\) and the corresponding amount of selected data at epoch \(t\).}
    \label{fig:curve}
\end{figure}

\citet{platanios2019competence} introduced the concept of ``Competence-Based Curriculum Learning'', highlighting that competence reflects the model's learning ability, which progressively increases from an initial value \(\lambda_{\text{init}}\) to 1 over a duration \(T\). At the $t$-th epoch, the model selects only those training data that are well-aligned with its current capabilities, defined by the condition \(d^c < p(t)\), where $p(t)$ is the model's learning competence. The curve $p(t)$ is depicted as the red curve in Figure \ref{fig:curve} and is computed using the following formula:

\begin{equation}
p(t) =
\begin{cases} 
\sqrt{\frac{t}{T} \left( 1 - \lambda_{\text{init}}^2 \right) + \lambda_{\text{init}}^2} & \text{if } t \leq T, \\
1.0 & \text{otherwise}.
\end{cases}
\end{equation}

When \(p(t) \geq 1.0\), the model selects 100\% of the training dataset.

\subsection{Aspect-Enhance Denoising Module (AED)}
This module enhances text-image alignment for sentiment analysis by using an aspect-guided attention mechanism to filter out irrelevant visual data and focus on extracting meaningful features tied to each aspect.
\subsubsection{Aspect-Based Enhance Sentic Attention(AESA)}
We leverages the Aspect-Aware Attention(A3M) Module from \cite{zhou2023aom} to filter out visual block noise—visual blocks that are very few or nearly irrelevant to the aspect.
A3M uses an aspect-guided attention mechanism as described by the following formula:
\begin{align}
    Z_t &= \tanh((W_{CA}H^{CA} + b_{CA}) \oplus (W_H h_t + b_H)), \\
    \alpha_t &= \text{softmax}(W_{\alpha}Z_t + b_{\alpha}),
\end{align}
where $H^{CA} = \{h^{CA}_1, h^{CA}_2, \dots, h^{CA}_n\}$ is the list of all $n$ noun in sentence, \(Z_t\) is the comprehensive feature extracted from both the noun list $H^{CA}$ and the hidden states $h_t$. 
 \(W_{CA}\), \(W_H\), \(W_{\alpha}\), \(b_{CA}\), \(b_H\), and \( b_{\alpha} \) are the learned parameters. \( \oplus \) is an concatenate operator. We then get the aspect-related hidden feature \( h^A_t \) by calculating the weighted sum of all candidate aspects following the equation:
\begin{equation}
    h^A_t = \sum_{i=1}^{k} \alpha_{t,i} h^{CA}_i.
\end{equation}
To mitigate noisy visual blocks, the parameter \(\beta_t\) is learned to aggregate the atomic feature \(h_t\) with its aspect-related hidden feature \(h_A^t\).
\begin{align}
    \beta_t &= \text{sigmoid}(W_{\beta}[W_1 h_t; W_2 h_A^t] + b_{\beta}), \\
    \hat{h}_t &= \beta_t h_t + (1 - \beta_t) h_A^t,
\end{align}
where $W_{\beta}$, $W_1$, $W_2$, and $b_{\beta}$ are parameters, and $[;]$ denotes the concatenation operator for vectors. $\hat{h}_t$ is the final output of A3M after the semantic alignment and noise reduction procedure.

We utilize SenticNet \cite{cambria-etal-2016-senticnet}, an external affective commonsense knowledge base, to enhance sentiment feature representations for each concept. The affective values in SenticNet range from \([-1, 1]\), where values closer to \(1\) indicate a stronger positive sentiment. The attention output \(\hat{h}_t\) is further refined by incorporating these affective values from SenticNet as follows:
\begin{align}
s_i &= W_S \cdot \text{SenticNet}(w_i) + b_S, \\
h^S_i &= \hat{h}_i + s_i
\end{align}
where \(w_i\) is the word in the sentence, and \(W_S\) and \(b_S\) are learned parameters. 

\subsubsection{Weighted Association Matrix}
First, we use the Spacy library to create matrix \(D\) representing the dependency tree, where \(\text{D}_{ij}\) is the distance between the \(i\)-th word and the \(j\)-th word in the tree.

Next, we initialize a zero-weighted association matrix \(A\), \( A \in \mathbb{R}^{(m+n) \times (m+n)} \), where the image features range from 1 to \( m \), and the text features range from \( m+1 \) to \( m+n \).
We divide matrix \( A \) into 3 regions: \( A_\textit{image-image} \) contains all \(A_{ij}\) with \((i,j \leq m)\),
\( A_\textit{text-image} \) contains all \(A_{ij}\) with \((i < m < j)\) or \((j < m < i)\), 
and \( A_\textit{text-text} \) contains all \(A_{ij}\) with \((i,j > m)\). We fill the values for \(A\) as follows:
\begin{itemize}
    \item For \( A_\textit{image-image} \), we initialize the main diagonal with 1. \textbf{(I)}
    \item For \( A_\textit{text-image} \), to ensure aspect-oriented directionality:
    \begin{itemize}
        \item If the \(i\)-th feature is an aspect, we set \(A_{ik} = \text{cos}(\hat{h}_i, \hat{h}_k)\) for \(0 \leq k \leq m+n\).
        \item Similarly, if the \(j\)-th feature is an aspect, we set \(A_{kj} = \text{cos}(\hat{h}_k, \hat{h}_j)\) for \(0 \leq k \leq m+n\). \textbf{(II)}
    \end{itemize}
    \item For \( A_\textit{text-text} \), we set \(A_{ij} = \text{cos}(\hat{h}_i, \hat{h}_j)\) if \(\text{D}_{ij} \leq \textit{threshold}\). In this paper, we set the \textit{threshold} to 2. \textbf{(III)}
\end{itemize}

The above conditions can be rewritten as follows:
\begin{align}
A_{ij} = 
\begin{cases} 
1 & \text{(I)}, \\
\text{cos}(\hat{h}_i, \hat{h}_j) & \text{(II) and (III)}, \\
0 & \text{otherwise}.
\end{cases}
\end{align}
where \(\text{cos}(\cdot)\) is the cosine function.

\subsubsection{Graph Convolutional Network (GCN)}

Based on the weighted association matrix $A$ above and the enhanced sentiment feature \(h^S_i\), we feed the graph into the GCN layers to learn the affective dependencies for the given aspect. Then each node in the l-th GCN layer is updated according to the following equation:
\begin{align}
    h^S_{i,0} &= h^S_i, \\
    h^S_{i,l} &= \text{ReLU}\left(\sum_{j=1}^{n} A_{ij} W_l h^S_{i,l-1} + b_l \right).
\end{align}
where \(h^S_{i,l}\) is the hidden state of i-th node at l-th GCN layer, $W_l$, $b_l$ are learned parameters.
\begin{table}[!t]
\centering
\caption{Statistics of two benchmark datasets}
\label{tab:dataset_statistics}
\resizebox{0.4\textwidth}{!}{%
\begin{tabular}{l|cccc}
\hline
\textbf{Datasets} & \textbf{Positive} & \textbf{Neutral} & \textbf{Negative} \\ \hline
Twit15 Train & 928 & 1883 & 368 \\ \hline
Twit15 Dev   & 303 & 670  & 149 \\ \hline
Twit15 Test  & 317 & 607  & 113  \\ \hline
Twit17 Train & 1508 & 1638 & 416  \\ \hline
Twit17 Dev   & 515  & 517  & 144 \\ 
\hline
Twit17 Test  & 493  & 573  & 168  \\ \hline
\end{tabular}%
}

\end{table}
\begin{table*}[t!]
\centering
\caption{Results of different approaches for JMASA task, Italic value denote for second-best result and bold-typed value for best result. The $\Delta$ values show the difference between our model and the previous state-of-the-art.}
\label{tab:jmasa-results}
\resizebox{0.9\textwidth}{!}{%
\begin{tabular}{c|c|c|c|c|c|c|c}
\hline
\textbf{Modality} & \textbf{Approaches} & \textbf{2015\_P} & \textbf{2015\_R} & \textbf{2015\_F1} & \textbf{2017\_P} & \textbf{2017\_R} & \textbf{2017\_F1} \\
\hline
&SpanABSA \cite{hu2019open}          & 53.7  & 53.9  & 53.8  & 59.6  & 61.7  & 60.6  \\
&D-GCN \cite{chen2020joint}             & 58.3  & 58.8  & 59.4  & 64.2  & 64.1  & 64.1  \\
\textbf{TEXT}&GPT-2 \cite{radford2019language}           & 66.6  & 60.9  & 63.6  & 55.3  & 59.6  & 57.4  \\
&RoBERTa  \cite{liu2019roberta}         & 62.4  & 64.5  & 63.4  & 65.3  & 66.6  & 65.9  \\
&BART \cite{yan2021unified}             & 62.9  & 65.0  & 63.9  & 65.2  & 65.6  & 65.4  \\
\hline
&UMT-collapse \cite{yu2020improving}     & 60.4  & 61.6  & 61.0  & 60.0  & 61.7  & 60.8  \\
&OSCGA-collapse \cite{wu2020multimodalrepresentation}    & 63.1  & 63.7  & 63.2  & 63.5  & 63.5  & 63.5  \\
&RpBERT-collapse \cite{sun2021rpbert}   & 49.3  & 46.9  & 48.0  & 57.0  & 55.4  & 56.2  \\
&CLIP \cite{radford2021learning}             & 44.9  & 47.1  & 45.9  & 51.8  & 54.2  & 53.0  \\
\textbf{MULTIMODAL}&RDS \cite{xu2022different}             & 60.8  & 61.7  & 61.2  & 61.8  & 62.9  & 62.3  \\
&JML* \cite{ju2021joint}             & 64.8  & 63.6  & 64.2  & 65.6  & 66.1  & 65.9  \\
&VLP-MABSA* \cite{ling2022vision}        & 64.1  & \textit{68.1}  & 66.1  & 65.8  & 67.9  & 66.9  \\
&AoM* \cite{zhou2023aom}           & \textit{65.15} & 67.6  & \textit{66.35} & \textit{65.94} & \textit{68.0}  & \textit{67.06} \\
&\textbf{\mName{} (Ours)}         & \textbf{66.1} & \textbf{68.18} & \textbf{67.1} & \textbf{66.35} & \textbf{68.2} & \textbf{67.3} \\
\hline
&$\Delta$  & \textbf{0.95} & \textbf{0.08} & \textbf{0.75} & \textbf{0.41} & \textbf{0.2} & \textbf{0.24} \\
\hline
\end{tabular}%
}
\end{table*}
\subsubsection{Prediction and Loss Function}

Based on \cite{lewis2019bart}, the BART decoder predicts the token probability distribution using the following approach:
\begin{align}
     \tilde{H} &= \alpha_1 \hat{H} + \alpha_2 \hat{H}^S ,\\
    h_t^d &= \text{Decoder}(\tilde{H}; Y_{<t}), \\
    \overline{H}_T &= \frac{W + \tilde{H}_T}{2}, \\
    P(y_t) &= \text{softmax}([\overline{H}_T; C^d] h_t^d), \\
    L &= -\mathbb{E}_{X \sim D} \left[ \sum_{t=1}^{O} \log P(y_t \mid Y_{<t}, X) \right],
\end{align}
where \(\hat{H}\) denote the output from the AESA module, and \(\hat{H}^S\) represent the output from the GCN. The parameters \(\alpha_1\) and \(\alpha_2\) indicate the respective contributions of \(\hat{H}\) and \(\hat{H}^S\). The hidden state of the decoder at time step \(t\) is denoted by \(h_t^d\). The term \(\tilde{H}_T\) refers to the textual portion of \(\tilde{H}\). The matrix \(W\) represents the embeddings for input tokens, and \(C^d\) denotes the embeddings for sentiment categories, $L$ is the loss function, $O = 2M + 2N + 2$ is the length of $Y$, and $X$ denotes the multimodal input.

\section{Experiment}
\subsection{Experimental Settings}
\textbf{Datasets}:In this study, we utilize two primary benchmark datasets: Twitter2015 and Twitter2017, as detailed by \cite{yu2019adapting}. The statistics of these two
datasets are presented in Table \ref{tab:dataset_statistics}.

\textbf{Evaluation Metrics}:
The performance of our model is assessed across different tasks using various metrics. For the MABSA and MATE tasks, we utilize the F1 score, Precision (P), and Recall (R) to evaluate the performance, and in the MASC task, we only adopt Accuracy (ACC) and F1 score.

\subsection{Comparison models}
We compare our model with all competitive baseline models list below:

\textbf{For JMASA Task:}
 SpanABSA \cite{hu2019open}, D-GCN \cite{chen2020joint}, GPT-2 \cite{radford2019language}, RoBERTa \cite{liu2019roberta}, BART \cite{yan2021unified}, UMT-collapsed \cite{yu2020improving}, OSCGA-collapsed \cite{wu2020multimodalrepresentation}, and RpBERT-collapsed \cite{sun2021rpbert}, CLIP \cite{radford2021learning}, RDS \cite{xu2022different}, JML \cite{ju2021joint}, VLP-MABSA \cite{ling2022vision}, AoM \cite{zhou2023aom}.
\textbf{For MASC Task:}
ESAFN \cite{yu2019entity}, TomBERT \cite{yu2019adapting}, CapTrBERT \cite{khan2021exploiting}. 
\textbf{For MATE Task:}
RAN \cite{wu2020multimodal}, UMT \cite{yu2020improving}, OSCGA \cite{wu2020multimodalrepresentation} 

\subsection{Main Results}
Table \ref{tab:jmasa-results} summarizes the results for the JMASA task. Our model achieves the highest scores across Precision, Recall, and F1 metrics on both the Twitter2015 and Twitter2017 datasets, with notable improvements of 0.95\%, 0.08\%, and 0.75\% in Precision, Recall, and F1 on Twitter2015, and 0.41\%, 0.20\%, and 0.24\% on Twitter2017 compared to the second-best results. This consistent performance across datasets demonstrates robust generalizability. For the MASC task as shown in Table \ref{tab:MASC}, our model shows F1 score increases of 0.63 and 0.34 on the Twitter2015 and Twitter2017 datasets, respectively, though accuracy metrics vary slightly. In the MATE task in Table \ref{tab:MATE}, F1 scores increase by 0.08 and 0.15 on Twitter2015 and Twitter2017, respectively, but there are inconsistencies in precision and recall metrics across datasets.



\begin{table}[h!]
\centering
\caption{Results of the MASC Task. Italicized values represent the second-best results, while bolded values indicate the best results. The $\Delta$ values denote the difference between our model and the previous SOTA model.}
\label{tab:MASC}
\resizebox{0.48\textwidth}{!}{%
\begin{tabular}{c|c|c|c|c}
\hline
\textbf{Methods} & \textbf{2015\_ACC} & \textbf{2015\_F1} & \textbf{2017\_ACC} & \textbf{2017\_F1} \\
\hline
ESAFN        & 73.4   & 67.4   & 67.8   & 64.2   \\
TomBERT      & 77.2   & 71.8   & 70.5   & 68.0   \\
CapTrBERT    & 78.0   & 73.2   & 72.3   & 70.2   \\
JML          & \textbf{78.7}   &         & 72.7   &          \\
VLP-MABSA    & 78.6   & 73.8   & \textit{73.8}   & 71.8   \\
AoM*         & 78.2   & \textit{73.81}  & 73.6   & \textit{72.05}  \\
\textbf{\mName{} (Ours)}    & \textit{78.62}  & \textbf{74.44}  & \textbf{74.14}  & \textbf{72.39}  \\
\hline
$\Delta$  & \textbf{-0.08} & \textbf{0.63} & \textbf{0.34} & \textbf{0.34}\\
\hline
\end{tabular}%
}

\end{table}

\begin{table}[h!]
\centering
\caption{Results of different approaches for MATE task, Italic value denote for second-best result and bold-typed value for best result. The $\Delta$ values denote the difference between our model and the previous SOTA model.}
\label{tab:MATE}
\resizebox{0.48\textwidth}{!}{%
\begin{tabular}{l|c|c|c|c|c|c}
\hline
\textbf{Methods} & \textbf{2015\_P} & \textbf{2015\_R} & \textbf{2015\_F1} & \textbf{2017\_P} & \textbf{2017\_R} & \textbf{2017\_F1} \\
\hline
RAN*           & 80.5  & 81.5  & 81.0  & 90.7  & 90.7  & 90.0  \\
UMT*           & 77.8  & 81.7  & 79.7  & 86.7  & 86.8  & 86.7  \\
OSCGA*         & 81.7  & 82.1  & 81.9  & 90.2  & 90.7  & 90.4  \\
JML*           & 83.6  & 81.2  & 82.4  & \textbf{92.0}  & 90.7  & 91.4  \\
VLP-MABSA*     & 83.6  & \textbf{87.9}  & \textit{85.7}  & 90.8  & 92.6  & \textit{91.7}  \\
AoM*           & \textit{83.72} & 86.79 & 85.23 & 89.58 & \textbf{92.71} & 91.12 \\
\textbf{\mName{} (Ours)} & \textbf{84.34} & \textit{87.27} & \textbf{85.78} & \textit{91.01} & \textbf{92.71} & \textbf{91.85} \\
\hline
$\Delta$  & \textbf{0.62} & \textbf{-0.63} & \textbf{0.08} & \textbf{-0.99} & \textbf{0.0} & \textbf{0.15} \\
\hline
\end{tabular}%
}

\end{table}

\subsection{Ablation Study}
\subsubsection{Module Effectiveness}
In this section, we evaluate the impact of each module on the model’s performance, as detailed in Table \ref{ablation-module}. Removing the Aspect-based Emotion Sentiment Analysis (AESA) module results in the most significant drop in performance, highlighting its crucial role in aspect alignment and the integration of external affective commonsense knowledge. The removal of the Hybrid Curriculum Denoising (HCD) module also leads to a substantial performance decrease, underscoring its importance in enhancing overall model effectiveness. On the other hand, omitting the Graph Convolutional Network (GCN) causes only a modest reduction in the F1-score compared to HCD, suggesting that while GCN is important for handling semantic and structural aspects of the data, its impact is less pronounced than that of HCD.
\begin{table}[h!]
\centering
\caption{Ablation Modules Performance}
\label{ablation-module}
\resizebox{0.45\textwidth}{!}{%
\begin{tabular}{c|c|c|c}
\hline
\textbf{Methods} & \textbf{2015\_P} & \textbf{2015\_R} & \textbf{2015\_F1} \\
\hline
\textbf{w/o AESA} & 62.5 & 62.7 & 62.6 \\
\textbf{w/o HCD} & 65.4 & 66.96 & 66.17\\
\textbf{w/o GCN} & 65.2 & 67.5 & 66.33 \\
\textbf{\mName{} (Ours)} & \textbf{66.1} & \textbf{68.18} & \textbf{67.1} \\
\hline
\end{tabular}%
}

\end{table}

\subsubsection{Ratio Contribution Test}
\begin{figure}[!ht]
    \centering
\includegraphics[width=1\linewidth]{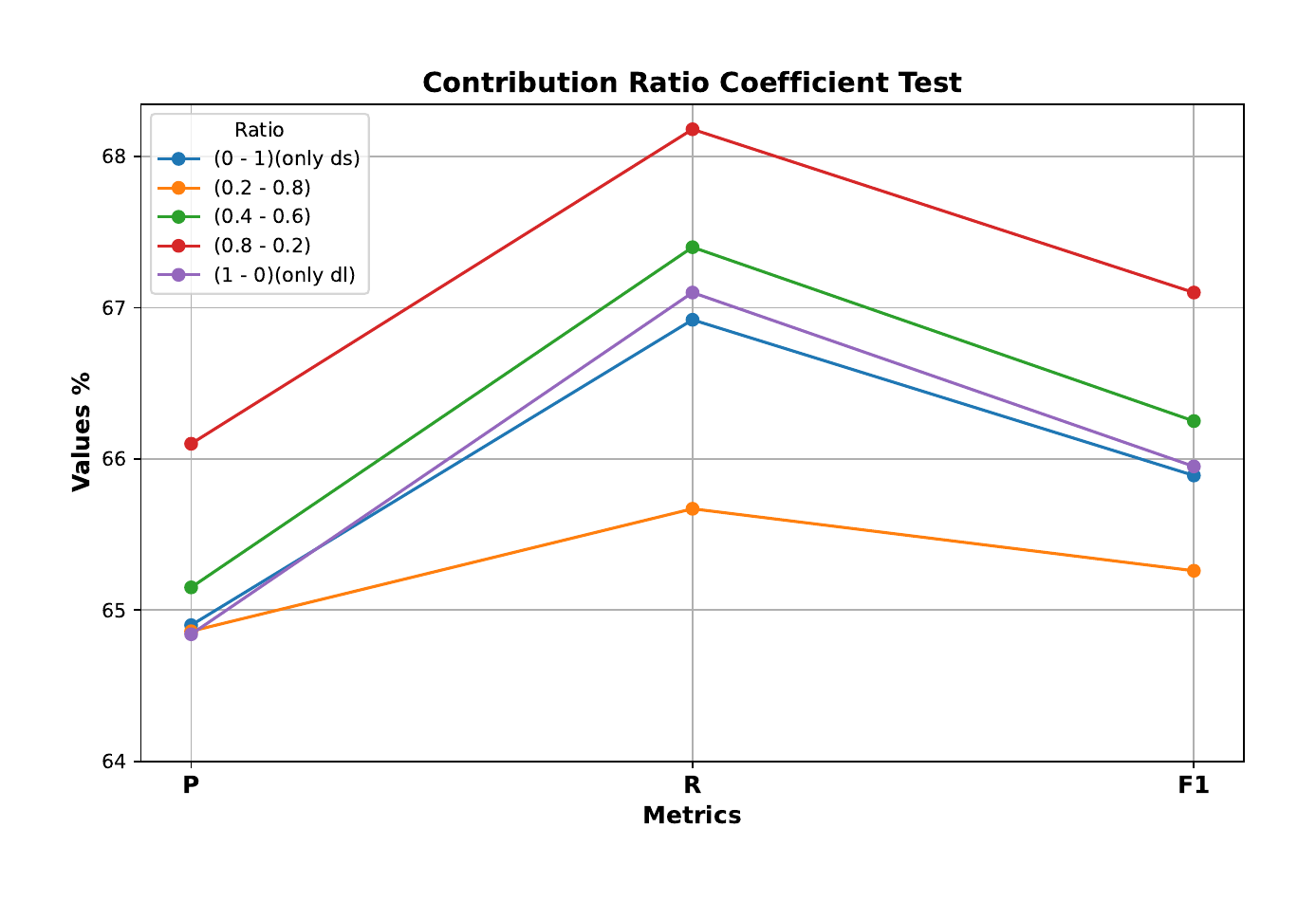}
    \caption{Illustration Contribution Ratio Coefficient Test}
    \label{fig:ratio}
\end{figure}

Figure \ref{fig:ratio} provides a detailed examination of the fine-tuning process for the contribution ratio of \(d_l\) - \(d_s\) at Hybrid Curriculum Denoising module (HCD), aiming to determine the optimal ratio for the model.
Based on Figure \ref{fig:ratio}, the ratio of (0.8 - 0.2) achieves the highest F1-score of 67.1. This indicates that the (0.8 - 0.2) ratio is the most effective configuration for optimizing model performance between \(d_l\) and \(d_s\). Therefore, we select this ratio as the optimal setting for the model.

\begin{figure*}[!t]
    \centering
    \includegraphics[width=0.65\linewidth]{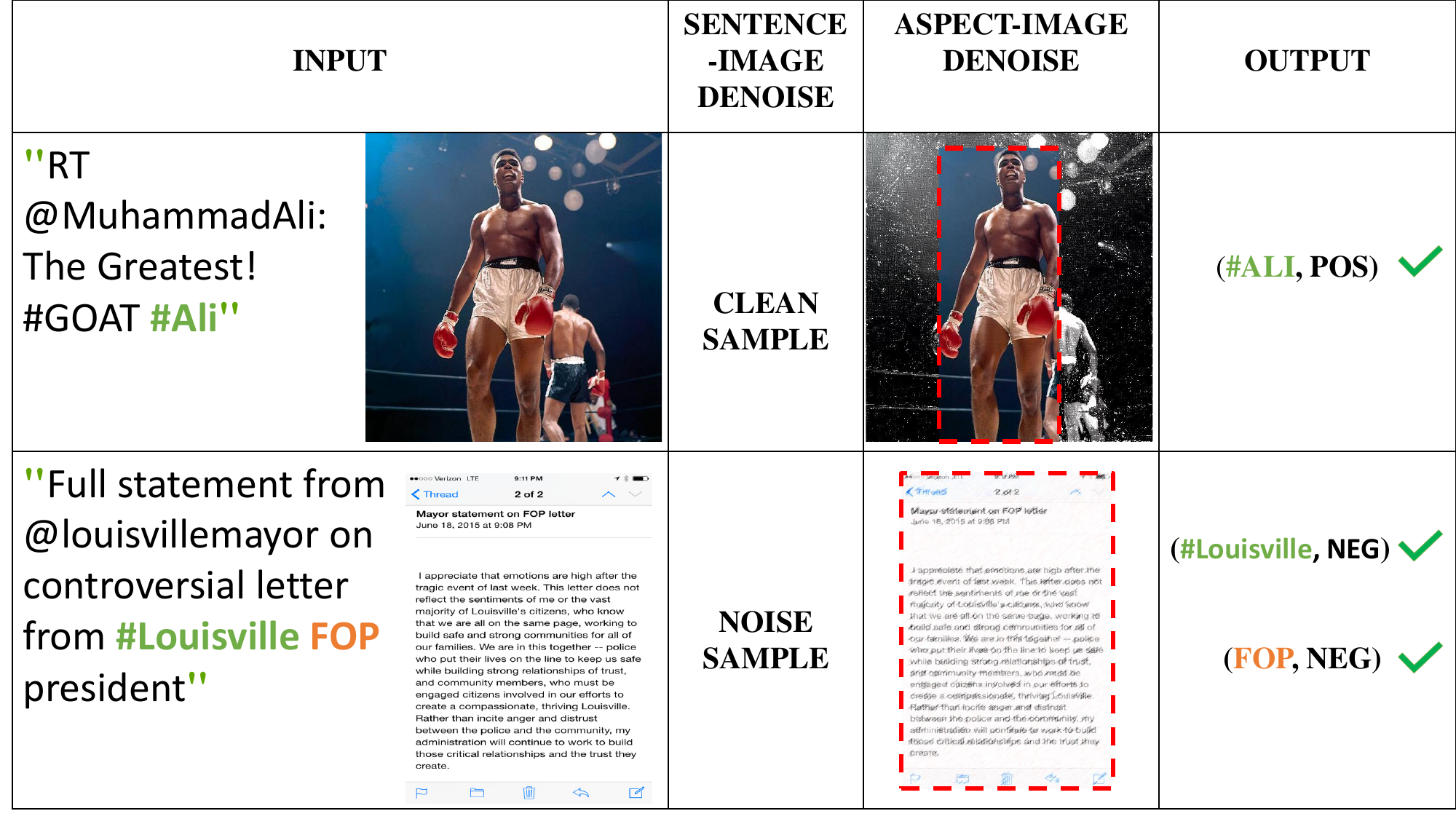}
    \caption{The figure illustrates instances where sentence-image noise and aspect-image noise impact the effectiveness of sentiment analysis. The easy sample features a clear alignment between the sentence and image, enhancing sentiment detection, while the hard sample involves a blurry image with minimal relevance to the sentence's aspects, complicating accurate sentiment evaluation.}
    \label{fig:case}
\end{figure*}
\subsection{Case Study}
Figure \ref{fig:case} illustrates how each module in our model processes data samples with two different levels of difficulty: ``easy'' (sample 1) and ``hard'' (sample 2). In the \textit{Sentence-Image Denoise} step, sample 1 is considered ``clean'' because the image is strongly related to the text, whereas sample 2 is not. In the \textit{Aspect-Image Denoise} step, the most important image regions related to the specific aspect are highlighted, while the blurred parts are considered noise and are not emphasized during training. The output represents the model's predictions for each sample, demonstrating the effectiveness of our model.

\section{Conclusion}
This paper introduced \mName{}, a novel framework for enhancing Multimodal Aspect-Based Sentiment Analysis (MABSA) by addressing both sentence-image and aspect-image noise. The framework comprises the Hybrid Curriculum Denoising Module(HCD), which utilizes Curriculum Learning to incrementally manage noisy data, and the Aspect-Enhanced Denoising Module(AED), which employs aspect-guided attention to filter irrelevant visual information. Empirical evaluations on the Twitter2015 and Twitter2017 datasets demonstrate that DualDenoise significantly improves Precision, Recall, and F1 scores compared to existing methods. These results affirm the model's efficacy in managing multimodal noise and its robust performance across diverse datasets. Future research may focus on refining the curriculum learning strategy and exploring broader applications of the proposed methodology.

\bibliography{custom}




\end{document}